\documentclass[11pt]{article}
\usepackage{coling2018}
\usepackage{times}
\usepackage{url}
\usepackage{latexsym}
\usepackage{tikz-qtree}
\usepackage{float}
\usepackage{comment}
\usepackage{multirow}

\usepackage{amsfonts}
\usepackage{amsthm}
\usepackage{amsmath}
\usepackage{paralist}

\newcounter{algocf}
\newcommand{\algoinout}[5]{
  \refstepcounter{algocf}
  \label{#1}
  \noindent
  \rule{\columnwidth}{0.025cm} \\
  {\bf Algorithm \thealgocf: #2} \\[-0.2cm]
  \rule{\columnwidth}{0.025cm} \\[-2\topsep]
  \begin{description}
  \item[Input.] #3
  \item[Output.] #4
  \end{description}
  \vspace{-1\topsep}
  \begin{enumerate}
    \renewcommand{\labelenumi}{\textbf{\theenumi}.}
    \renewcommand{\labelenumii}{\theenumii.}
    \renewcommand{\labelenumiii}{\theenumiii.}
    \renewcommand{\labelenumiv}{\theenumiv.}
  
    \renewcommand{\theenumi}{Step~\arabic{enumi}}
    \renewcommand{\theenumii}{Step~\arabic{enumi}.\arabic{enumii}}
    \renewcommand{\theenumiii}{Step~\arabic{enumi}.\arabic{enumii}.\arabic{enumiii}}
    \renewcommand{\theenumiv}{Step~\arabic{enumi}.\arabic{enumii}.\arabic{enumiii}.\arabic{enumiv}}
  
    \setlength{\itemsep}{2ex}
    \setlength{\parskip}{0pt}
    \setlength{\parsep}{0pt}
    #5
  \end{enumerate}
  \vspace{-0.5cm}
  \rule{\columnwidth}{0.025cm}
}

\newcommand{\tuple}[1]{\ensuremath\left\langle{#1}\rangle\right}

\title{Semantic Parsing: Syntactic assurance to target sentence\\ 
using LSTM Encoder CFG-Decoder}

\author{Fabiano Ferreira Luz {\normalfont and} Marcelo Finger\\
University of Sao Paulo - USP\\
Institute of Mathematics and Statistics - IME\\
{\tt fluz,mfinger@ime.usp.br}}

\date{}

\begin{document}
\maketitle
\begin{abstract}
Semantic parsing can be defined as the process of mapping natural
language sentences into a machine interpretable, formal representation
of its meaning. Semantic parsing using LSTM encoder-decoder neural
networks have become promising approach. However, human automated
translation of natural language does not provide grammaticality
guarantees for the sentences generate such a guarantee is particularly
important for practical cases where a data base query can cause
critical errors if the sentence is ungrammatical. In this work, we
propose an neural architecture called Encoder CFG-Decoder, whose
output conforms to a given context-free grammar. Results are show for
any implementation of such architecture display its correctness and
providing benchmark accuracy levels better than the literature.

\end{abstract}

\section{Introduction}
\label{intro}

\noindent 
Semantic Parsing can be defined as the process of mapping natural language sentences into a machine interpretable, formal representation of its meanings. Currently, there are many efforts aiming at transforming human language into a computational representation \cite{zettlemoyer2005learning,alshawi2014deterministic,bowman2014recursive}.
Some of this approaches are basead on neural network machine translation techniques which however do not guarantee that output complies with machine language syntactical form.

This work focuses on producing queries to ontology using SPARQL query language, using a neural model for semantic parsing that ensures that the output text obeys a given context-free grammar (CFG).  For that, we propose an enrichment of the usual encoder-decoder neural model used in natural language translation, which we call the \emph{encoder CFG-decoder}, that ensures both learning and prediction that the generated sentence conforms to a given CFG.

We describe experimental results on two datasets: \textsf{Geo880}\footnote{\url{http://www.cs.utexas.edu/users/ml/nldata/geoquery.html}.}, a set of 880 queries to a database of United States geography; and \textsf{Jobs640}\footnote{\url{ftp://ftp.cs.utexas.edu/pub/mooney/nl-ilp-data/jobsystem/}.}, a set of 640 queries to a database of job listings. Previous works in the literature have also used these same datasets~\cite{tang2001using,zettlemoyer2005learning,dong2016language}; howerver, none of those semantic parsers had SPARQL as a target language. In our case, we have developed a SPARQL version of each natural language query in the dataset. For that reason, we do not make a direct comparison with these previous works but, instead, we built semantic parsers into SPARQL using known implementations of neural networks with the encoder-decoder architecture. Our method achieves 83.25\% of accuracy to \textsf{Geo880} and 85.71\% of accuracy to \textsf{Jobs640}, outperforming the accuracy obtained from the known implementations of encoder-decoder neural networks.

This work is organized as follows. Section~\ref{sec:related} presents some related work. A brief, but enough, description of essential concepts to understand our proposal is shown in Section~\ref{sec:background}. In Section~\ref{sec:sats}, we talk about the syntactic guarantee that we included in the Encoder CFG-Decoder model. Finally, the experimental evaluation of the proposed model and the discussion and future works are presented at Section~\ref{sec:experimental} and~\ref{sec:discussion}, respectively.

\section{Related Word}
\label{sec:related}

Some important works address the use of the encoder-decoder neural network architecture~\cite{cho2014learning} as a natural language translation model  whose target is a  formal language. Among those we can mention~\cite{dong2016language}, which have proposed a semantic parser based on LSTM encoder-decoder architecture and developed the hierarchical tree-decoder.  The translation method is based on an attention-enhanced sequence-to-sequence model. The input sentence is encoded into vector representations using recurrent neural networks, and generate their logical forms by conditioning the output on those encoding vectors. The model is trained in an end-to-end fashion to maximize the likelihood of target logical forms given the natural language inputs. In the same work the authors introduce a more powerful approach from decoder to formal language. However those approaches produce output that still lacks grammatical guarantees.

The task of natural language translation into SPARQL is an important one given that DBpedia \cite{auer2007dbpedia}, currently possessing thousands of data items, can be queried by using SPARQL; furthermore,m the  growth of linked data cloud provides further data items to be required using SPARQL.  The work of \cite{soru2017sparql} aims at translating natural language to SPARQL and employs automated translation techniques, where SPARQL is considered a foreign language, without guarantees of generating grammatical SPARQL queries. That model is called ``Neural SPARQL Machines'', which is mainly focused in Question Answering on Liked Data. According to the authors, their model can work very well for the problems of vocabulary mismatch and perform graph pattern composition.

A previous implementation of a semantic parser having SPARQL as a target language was developed employing an encoder-decoder with neural attention~[Ommitted]; however, such an implementation could not guarantee the generation of syntactically correct queries.  In fact, 5.7\% of queries generated by that implementation contained syntactical errors.

\section{Background}
\label{sec:background}

\subsection{Recurrent Neural Network}

A Recurrent Neural Network (RNN) is a type of artificial neural network where connections between units form a directed cycle. This cycle represents an internal state of the network which allows it to exhibit dynamic temporal behavior. RNNs can use their internal memory to process arbitrary sequences of inputs.

It has been noted by~\cite{bengio1994learning} that RNNs suffer from the \textit{vanishing gradient problem}, which consists of the exponential decrease that the value of $h_{t'}$ has influence over the value of $h_t$, $t' < t$, leading to a very short temporal memory in the network. One solution to this problem was a change in the network architecture called Long Short Term Memory (LSTM)~\cite{hochreiter1997long}. RNN-LSTM has been used successfully in language modeling problems it can handle long sequences adequately.

\subsection{Encoder-Decoder Model}
\label{sec:enc-dec}
The encoder-decoder model proposed by \cite{cho2014learning} is a neural network architecture that learns the conditional distribution of a conditioning variable-length sequence $\mathbf{x}$ in another variable-length sequence $\mathbf{y}$.  It performs this task by learning how to \textit{encode} a variable-length sequence $x_1,...,x_{t}$ into a fixed-length vector representation $c$ and then to \textit{decode} a given fixed-length vector representation $c$ back into a variable-length sequence $y_1,...,y_{s}$. The function may be interpreted as the distribution $p(y_1,...,y_{s} | x_1,...,x_{t})$, where the input sequence length $t$ and output one $s$ can be different.

The \emph{encoder} is an RNN that reads each word of an input sequence $x_1,...,x_{t}$ sequentially. As it reads each symbol, the hidden state $h_t$ of the RNN is updated according to the equation $h_t=f(h_{t-1},x_t)$, where $h_{t-1}$ is the value of the hidden layer at time $t-1$, $x_t$ is the input feature vector, and $f(.)$ is a nonlinear function. After reading the end of the sequence (marked with an end sequence symbol), the hidden state of the RNN is encoded into $c$. In order to simplify we can define $c$ as the output $h$. 

\begin{figure}[H]
  \centering
  \includegraphics[width=.40\textwidth]{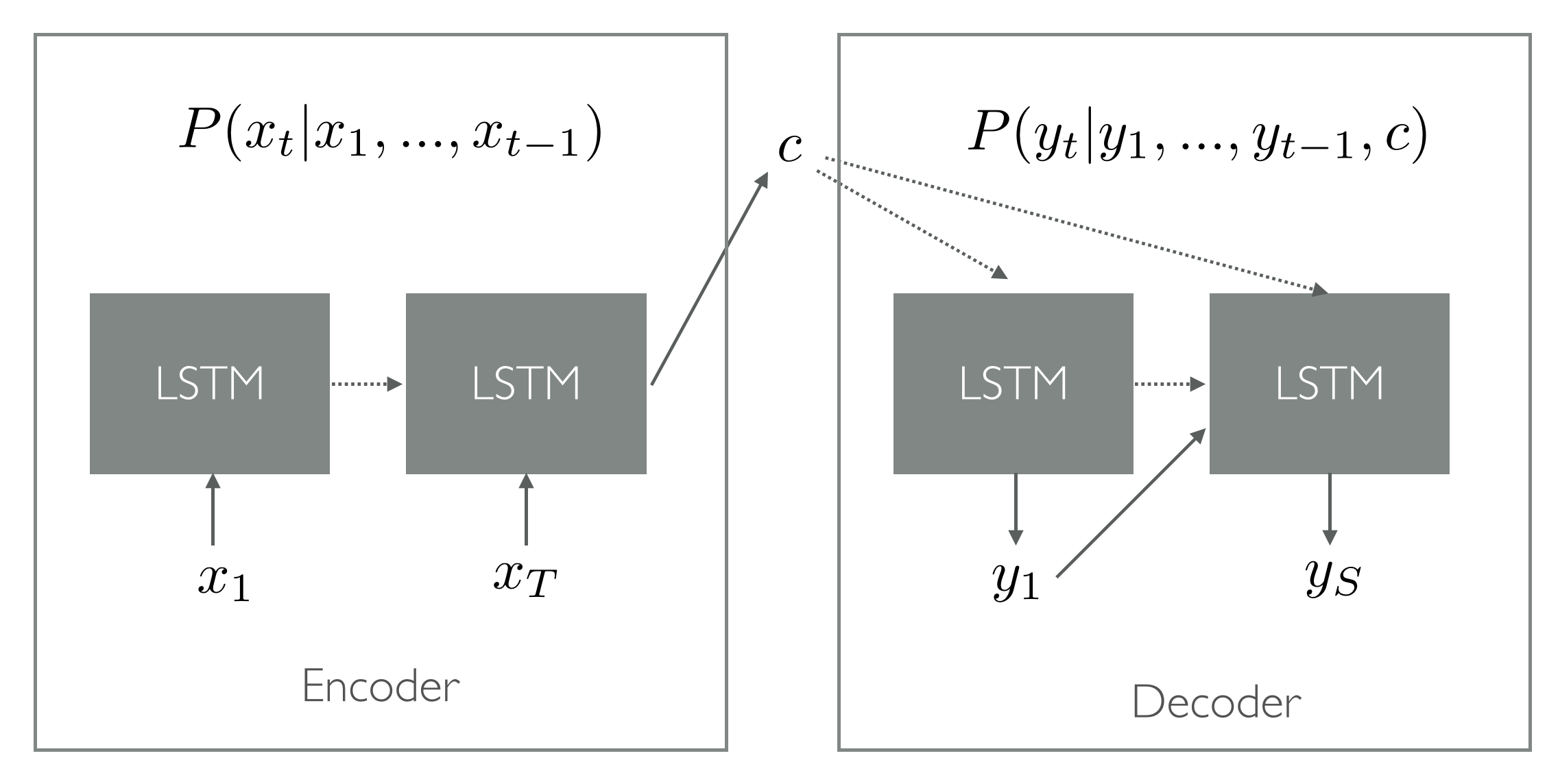} 
  \caption{encoder-decoder scheme. }
  \label{fig:enc-dec} 
\end{figure}

The \emph{decoder} is another RNN which is trained to generate the output sequence by predicting the next symbol $y_t$ given the hidden state $h_t$. However, unlike the RNN described previously, here both $y_t$ and $h_t$ are conditioned to $y_{t-1} $ and the encoding $c$ of the input sequence. Thus, the hidden state of the decoder at time $t$ is computed by: $h_t=f(h_{t-1},y_{t-1},c)$, and likewise, we can define the conditional distribution of the next symbol by the following equation:
\begin{align}
\label{eq:prob-dec}
P(y_t|y_{t-1},...,y_1,c)=g(h_{t-1},y_{t},c).
\end{align}
The activation function $g$ produces valid probabilities by, for example, computing the \textit{softmax function}. Figure~\ref{fig:enc-dec} presents an overview of the encoder and decoder scheme.

The combination of the two described components (Encoder and Decoder) make up the training of the proposed model to maximize the conditional log-likelihood and can be represented by the equation:
\vspace{-1ex}
\begin{align}
\label{eq:train}
\max_{\theta} \frac{1}{N} \sum\limits_{n=1}^{N} \log p_{\theta}(\mathbf{y} | \mathbf{x}),
\end{align}
where $\theta$ is the set of the model parameters and each pair $(\mathbf{x},\mathbf{y})$
is, respectively, an input and output sequence. In our case, we use the vector representation of questions in natural language as input and the SPARQL query as an output. As the output of the decoder, starting from the input, is differentiable, we can use a gradient-based algorithm to estimate the model parameters.

After training the LSTM encoder-decoder, the model can be used in two distinct ways. In the first case, we can use the model to generate a target sequence, once the input sequence is provided computing the most probable output given the input. In the second one, the model can be used to evaluate a given pair of input and output sequences by calculating a score (e.g. the probability $p_\theta(\mathbf{y} | \mathbf{x})$). 

\subsection{SPARQL and OWL Ontologies}

In computing, ontologies are used to capture knowledge of a particular domain of interest. We can say that ontology is a formal specification of a conceptualization \cite{gruber1993}. \cite{borst1997} extends the definition when says that this conceptualization can also be shared.

The Web Ontology Language (OWL) is a formal language for the description of ontologies \cite{bechhofer2004} and was developed as an extension of the RDF vocabulary (Resource Description Framework). RDF defines the data structure and OWL describes the semantic relationship between them. Simple or complex concepts can be developed by using a wide range of operators that the OWL language offers.

SPARQL is a query language able to retrieve and manipulate data stored in the RDF format (Resource Description Framework) which is the basis of the OWL language. RDF is a labeled and directed data format that is used to represent information on the Web \cite{prud2008sparql}.

We employ SPARQL using the following features:

\begin{description}
  \setlength\itemsep{-1ex}
\item[Query Forms.] SPARQL has four query forms, but in the present work we used just two of them (SELECT, ASK). These query forms use the solutions from pattern matching to form result sets or RDF graphs \cite{prud2008sparql}.
\item[Graph update.] This operations change existing graphs in the Graph Store but do not explicitly delete nor create them \cite{gearon2012sparql}. In this project we used just (INSERT).
\item[RDF triples.] It is conventionally written following: subject, predicate, object. The predicate is also known as property of the triple \cite{klyne2006resource}. The subject and object are names for two "things" in the world, and the predicate is the name of a relationship between the two.
\item[Filters.] A constraint, expressed by the keyword FILTER, is a restriction on solutions over the whole group in which the filter appears \cite{prud2008sparql}.
\end{description}

\section{Syntactic Assurance to Target Sentences}
\label{sec:sats}

Our work presents a refinement of the encoder-decoder architecture. We are primarily concerned with generating a target language that obeys a given context-free grammar. We are also concerned with the construction of training corpora for the case where the target language must obey a given CFG.  

A CFG is a 4-tuple $\mathcal{G} = (\mathcal{V},\Sigma,\mathcal{R},S_0)$, where $V$ is a finite set of \emph{nonterminal} symbols; $\Sigma$ is the vocabulary, consisting of a finite set of \emph{terminal} symbols, $V \cap \Sigma = \emptyset$; $S_0 \in V$ is the starting nonterminal; and $\mathcal{R}$ is a set of CFG rules, which we consider here as partitioned into two sets of rules: the terminal rules are of the form $X \to t$, $X \in V$ and $t \in \Sigma$; and the nonterminal rules of the form $X \to \mathbf{Y}$, $X \in V$ and $\mathbf{Y} \in V^*$.

The parsing of an input sentence $x_1, \ldots x_n$ generates a \emph{parsing tree}, whose leaf is the input sentence and whose internal nodes are nonterminals.  We say that an occurrence of a nonterminal $X$ in a parsing tree dominates the sequence $x_t, \ldots, x_{t+r}$ if the subtree rooted in $X$ has leaves $x_t, \ldots, x_{t+r}$.  The initial symbol $S_0$ always dominates the whole input sentence.

\subsection{SPARQL Grammar}
\label{sec:grammar}
To illustrate our method we restrict ourself to a fragment of SPARQL queries given by the grammar below. The first three rules are the proper grammatical rules; the remaining ones are the lexical rules which generate the language nonterminals. In this example we analyze the query \texttt{SELECT ?area \{ ?capital p:area ?area . ?texas p:has\_capital ?capital \}}:

\begin{center}
{\small
\begin{tabular}{| l c l p{0.3cm} l c l p{0.3cm} l c l|}
\hline
S   & $\to$ & C BL CT BR & & BL  & $\to$ & \{      & & P  & $\to$ & p:area         \\
C   & $\to$ & SE VA      & & BR  & $\to$ & \}      & & O  & $\to$ & ?area          \\
CT  & $\to$ & T Dot T    & & SE  & $\to$ & SELECT  & & J  & $\to$ & ?texas         \\
T   & $\to$ & J P O      & & VA & $\to$ & ?area    & & P  & $\to$ & p:has\_capital \\
Dot & $\to$ & .          & & J  & $\to$ & ?capital & & O  & $\to$ & ?capital       \\ \hline
\end{tabular}
}
\end{center}

This grammar generate the following parsing tree:

\begin{figure}[H]
	\begin{scriptsize}
	\begin{center}
	\begin{tikzpicture}[grow=down]
\tikzset{level distance=0.6cm} 
	\Tree[.S [.C [.SE SELECT ] [.VA ?area ]  ] [.BL  \{ ] [.CT [.T   [.J ?capital ] [.P p:area ] [.O ?area ] ]  [.Dot . ] [.T   [.J ?texas ] [.P p:has\_capital ] [.O ?capital ] ] ] [.BR \} ] ] 
	\end{tikzpicture}
	\end{center}
	\end{scriptsize}
\end{figure}

\subsection{Encoder CFG-Decoder Model}

In our approach we add some elements to the traditional encoder-decoder model in order to improve the training and generation phases.  The method takes as input a natural language and a grammar and gives as output a query generated by that grammar.  It is important to notice that the query is generated backwards, starting from its rightmost symbol and progressing leftward to the first symbol.  In this way, the tail of a query is available as a ``context'' $\chi = y_{s-r+1}, \ldots, y_{s-1},y_s$ during query generation, and it will be used both for encoding and decoding.  In both training and generation, the context $\chi$ is initially empty and grows during the execution of the process.

The encoder CFG-decoder models consists of a set of pairs of encoder-decoder networks, one for each nonterminal, $\{\tuple{\mathcal{E}_{X},\mathcal{D}_{X}}| X \in \mathcal{V}\}$.  During the generation process, there is a controller mechanism that, given a context $\chi = y_{s-r+1}, \ldots, y_{s-1},y_s$, chooses a nonterminal $X \in \mathcal{V}$ for expansion, such that in the output parsing tree $X$ will dominate $y_u,\ldots,y_{s-r}$.  

The controller is simply a stack $\mathcal{S}$ of nonterminal symbols, and the choice of which nonterminal is expanded is obtained simply by popping the top nonterminal in $\mathcal{S}$, and the decoder can push new items to $\mathcal{S}$ according to the rules of the input grammar. Initially, $\mathcal{S}$ contains the single nonterminal $S_0$, the grammar initial nonterminal, and $\chi=\varepsilon$ the empty string. 

At each generation step, a nonterminal $X$ is popped from $\mathcal{S}$.  The encoder $\mathcal{E}_X$ receives as input the whole natural language sentence $W$ and the current context $\chi$, and generates the encoding $c$.  The decoder $\mathcal{D}_X$ then receives as input $c$ and the context $\chi$, and it has to choose from the grammar rules of the form $X \to \mathbf{Y}_i$ which one will be expanded.

There may be $k$ candidate rules headed by $X$, say, $X \to Y_{i1} Y_{i2} \ldots Y_{im_i} \bullet$, $1 \leq i \leq k$, where $\bullet$ is the stop element in the rule's tail. The decoder's task is to choose one specific rule tail among the possible one, by generating a string that maximizes the probability
\vspace{-1ex}
\begin{align}
\label{eq:prob-cfg-dec}
\rho_+ = P(Y_{it}|Y_{t-1},...,Y_1,\chi,c_X).
\end{align}

\begin{figure}[H]
  \begin{scriptsize}
    \begin{center}
      \begin{tikzpicture}[grow=down]
        \tikzset{level distance=0.6cm} 
	\Tree[.$S_1$ C BL [.$CT_3$ T  dot  [.$T_4$  J $P_6$ [.$O_5$ ?capital ]  ] ] [.$BR_2$ \} ] ]
      \end{tikzpicture}
    \end{center}
    \caption{Partial generation. The subindex in each symbol means the
      order each rule is expanded. In this illustration, the next rule
      to be expanded is the one headed by $P$. The input sentence and
      the context $\chi = \}$ ?capital is passed to the encoder
      $\mathcal{E}_p$ to generate the encoding $c$, that will be used
      next by the decoder~$\mathcal{D}_p$.}
    \label{fig:pg}
  \end{scriptsize}
\end{figure}
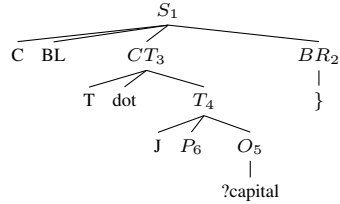

Each decoding step outputs a symbol $Y_j$, narrowing down which rule can be chosen. The main idea behind this is to eliminate all candidates whose $j$-th symbol is different from $Y_j$. At the end, a single rule is chosen. The strategy to choose this rule is as follows. Let $m = \max_{1\leq i\leq k} \{m_i\}$ and suppose we are in the $\ell$-th generation step. If $\ell>m$, we are done, since all rules are distinct and at each generation step $j$ we eliminate all the rules with $j$-th nonterminal symbol different from $Y_j$. Otherwise, given the probability $\rho$ of $Y_{j-1}$, if there is any rule with size $j-1$ and $\rho_+<\rho$, we choose the rule $X$ that finishes with $Y_{j-1}$ and we are done. If the rule chosen is a terminal $X \to t$, the context is updated by concatenating $t$ to the front of the current context.  If a nonterminal rule is chosen $X \to Y_1 \cdots Y_\ell$, the tail is pushed to the stack $\mathcal{S}$, so that its top now contains $Y_\ell$. The generation process then proceeds until $\mathcal{S}$ becomes empty. We illustrate this process in Figure~\ref{fig:pg}.


\subsubsection{The Training Algorithm}

During training we assume that we have a tree in which the leaves contains the target query. The decoder generates a tree on a top down, depth-first fashion. The actual query is generated right to left, that is, the last symbols of the output query sentence are produced first. This strategy allows us to consider the output sentence tail as a context used during query generation; that tail is also used in training of the encoder.

We perform the training phase using a supervised dataset that consists
on a list of pairs $(W,q)$, where $W$ is the English sentence and $q$ is
the corresponding SPARQL. Algorithm~\ref{alg:cfgt} describes the
training process and is called for each pair $\left\langle\textrm{sentence, query}\right\rangle$ of the training set for $\mathcal{E}$ and $\mathcal{D}$ weight readjustment.

\algoinout{alg:cfgt}{Encoder CFG-Decoder training}{A sentence $W$ and its corresponding SPARQL query $q$. A SPARQL context-free grammar $\mathcal{G} = (\mathcal{V},\Sigma,\mathcal{R},S_0)$. The context size $r$.}{The LSTM encoder-decoder trained model.}{\item \textbf{Initialization.} Generate the parser tree $\mathcal{G}_q = (\mathcal{V},\Sigma,\mathcal{R}_q,S_0)$ from SPARQL query $Q$. Let $X$ be the nonterminal under expansion, $X \gets S_0$; let $c$ the current encoding, $c \gets \varepsilon$; $\chi$ is the decoding context, $\chi \gets \varepsilon$. Consider an stack or nonterminals, $\mathcal{S} \subset \mathcal{V}_q$, initially empty.
\item \label{cfgt:train} \textbf{Training.}
  \begin{enumerate}
  \item Let $r \in \mathcal{R}_q$, $\mathcal{E}_X$, and $\mathcal{D}_X$ be the rule, the encoder, and the decoder associated with the nonterminal $X$, respectively.

  \item If $ = X \to Y_1 \cdots Y_m$ is a nonterminal rule,
    \begin{enumerate}
    \item Push $Y_1 \cdots Y_m$ to $\mathcal{S}$, in the same order they appear in $r$.
    \item Perform a forward step in $\mathcal{E}_X$ using the sentence $W$ and the context $\chi$, generating the encoding $c$.
    \item Perform a forward step in $\mathcal{D}_X$ using encoding $c$, unfolding its output $m$ times to obtain $Z_1 \cdots Z_m$; compute the error with respect to $Y_1 \cdots Y_m$ and store it for the backward step.
    \end{enumerate}
    
  \item If $r=X \to t$ is a terminal rule, concatenate $t$ to $\chi$. If $\chi$ has size greater than
    $r$, remove the rightmost character from $\chi$.

  \item Perform a backward step in $\mathcal{D}_X$ and $\mathcal{E}_X$. 
  \end{enumerate}
  
\item If $\mathcal{S}$ is not empty, let $X$ be its the top item; pop $X$ from $\mathcal{S}$ and go to
  \ref{cfgt:train}. Otherwise, return $\mathcal{E}$ and $\mathcal{D}$.
}

The main element used in the training phase is a \textit{parser tree} $\mathcal{G}_q$. We generate this parser
tree from $q$ using the SPARQL grammar we defined in Section~\ref{sec:grammar} and an Earley parser \cite{earley1970efficient}. This parser analyzes the query $q$ and generates $\mathcal{G}_q$ as a list of rules $\mathcal{R}_q$ used to generate $Q$.

%

\begin{figure}[H]
  \centering
  \includegraphics[width=.6\textwidth]{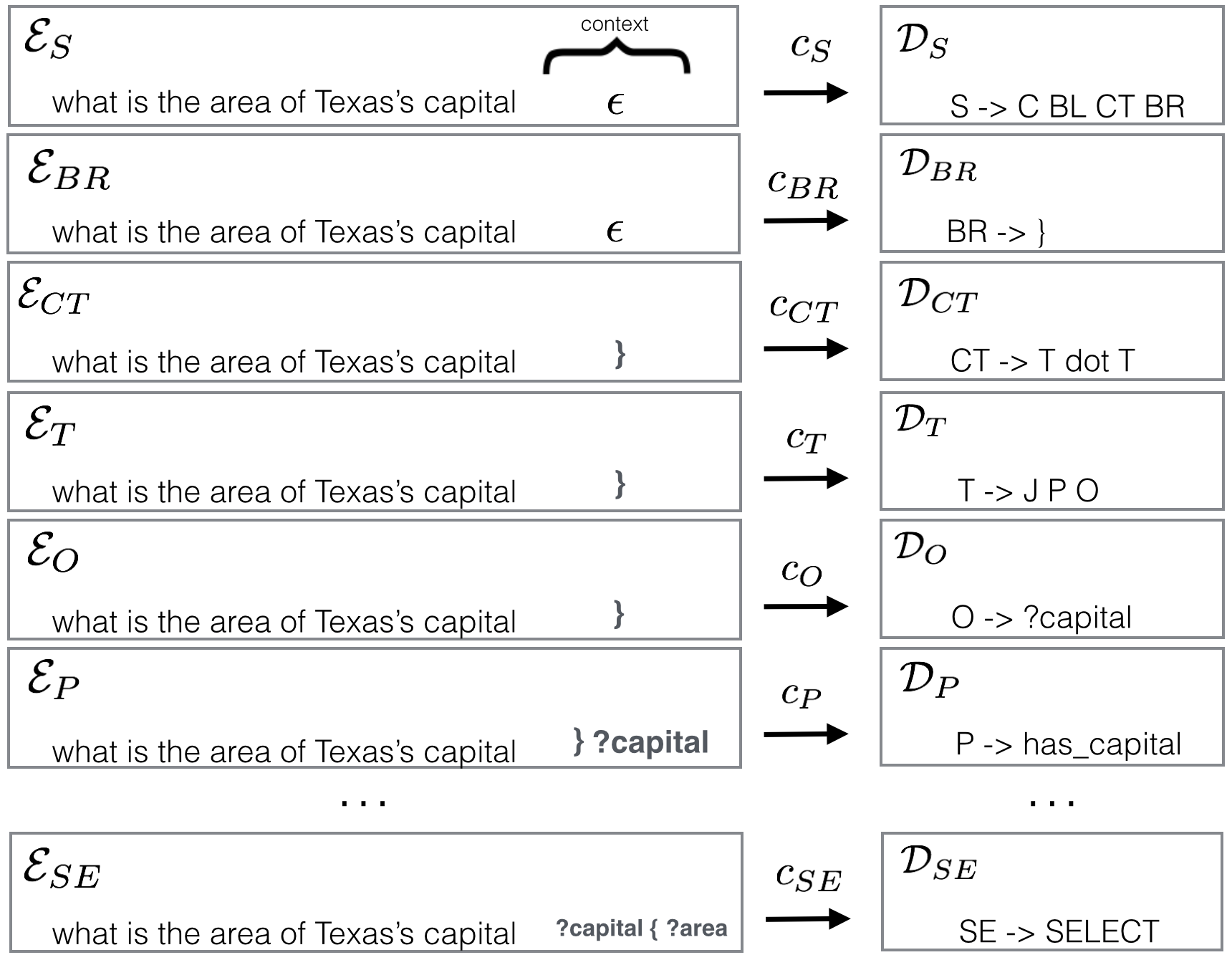} 
  \caption{CFG Decoder scheme (3-context).}
  \label{fig:multencdec}
\end{figure}

The encoder-decoder model consists of a set of $n$ encoders and $n$ decoders, where $n$ is the number of nonterminals in the SPARQL grammar. This means that we have a specialist encoder and decoder for each nonterminal symbol. 

In the query generation phase, as we allow for more than one rule with the same nonterminal at the head, we need a mechanism to choose the an appropriate rule for a given nonterminal.  For that, we added the notion of a \textit{context}, $\chi$, which provides the decoder with information a based on which a choice can be made.
The context consists of $s$ terminal symbols in the generated query; as the query is generated from right to left, it consists of the latest $s$ terminals generated. We illustrate this feature in Figure~\ref{fig:multencdec}, the context is the last terminal symbol from the right-hand side of the input SPARQL $q$, i.e. $s=3$.

In the Figure~\ref{fig:multencdec} we use the notation $c_X$, where $X \in V$ is a nonterminal symbol to indicate that each neural network generates a specialized $c$ encoding to $X \rightarrow \mathbf{Y}$ rule, where $X$ is a symbol of head and $\mathbf{Y}$ is a set of symbols that compose the tail.

\subsubsection{Generation}

The generation (also called prediction) process is decribed in Algorithm~\ref{alg:cfgp}. It followws the description of the runtime algorithm given above.

\algoinout{alg:cfgp}{Encoder CFG-Decoder generation}{A sentence $W$, an
  LSTM encoder-decoder trained model, and a SPARQL context-free grammar
  $\mathcal{G} = (\mathcal{V},\Sigma,\mathcal{R},S_0)$. The context size
  $\alpha$.}{A SPARQL query $q$.}{

\item Consider an empty stack $\mathcal{S} \subset \mathcal{V}$. Let $X$ be an empty string, the current nonterminal $X \gets S_0$, encoding $c \gets \varepsilon$ and context $\chi \gets \varepsilon$.

\item \label{cfgp:start} Let $\mathcal{R}_X \subset \mathcal{R}$,
  $\mathcal{E}_X$, and $\mathcal{D}_X$ the set of rules, the encoder and
  the decoder associated with the nonterminal $X$, respectively.

\item Compute an encoding $c$ using the encoder $\mathcal{E}_X$ with the
  sentence $W$ and the context $\chi$. Let $j$ be the latest position generated by the decoder, $j \gets 0$, and let $\rho$ be the probability of the symbol generated at $j$, $\rho \gets 0$.

  \begin{enumerate}

  \item Let $\ell$ be the size of the largest rule $r \in
    \mathcal{R}_X$. If $\ell < j$, go to \ref{cfgp:cont}. Otherwise, let
    $\mathcal{L}$ be a list containg the distinct $j$-th symbols from
    the rules in $\mathcal{R}_X$.

  \item Check which rule $X \in \mathcal{L}$ has the largest
    probability $\rho_+$ using the decoder $\mathcal{D}_X$ with the
    encoding $c$. Let $v$ be the $j$ symbol of $X$.

  \item If there is any rule in $\mathcal{R}_X$ with size $j-1$ and
    $\rho_+ < \rho$, remove all rule from $\mathcal{R}_X$ with size
    greater or equal than $j$ and go to \ref{cfgp:cont}.

  \item Remove from $\mathcal{R}_X$ the rules with size less than $j$
    and whose $j$-th symbol is different from~$v$.

  \item Let $\rho \gets \rho_+$, $j \gets j+1$, and go to Step 3.1.

  \end{enumerate}

\item \label{cfgp:cont} Let $r$ be the remaining rule in
  $\mathcal{R}_X$. If $r$ is a nonterminal rule, add to
  $\mathcal{S}_X$ the nonterminals from the tail of $r$, in the same
  order they appear in the rule. Otherwise, let $\sigma \in \Sigma$ be
  the symbol from the tail of $r$. Concatenate $\sigma$ in the
  beginning of $c$ and $Q$. If $c$ has size greater than $\alpha$,
  remove the rightmost character from $c$.
  
\item If $\mathcal{S}_X$ is not empty, let $p$ be its the top item;
  remove $p$ from $\mathcal{S}_X$ and go to
  \ref{cfgp:start}. Otherwise, return $Q$.
}


\section{Experimental Evaluation}
\label{sec:experimental}

We compared our methodology with related work using the two well-known datasets \textsf{Geo880} and \textsf{Jobs640}. Here we first define metrics for comparing different implementations and then describe an adaptation of the datasets to SPARQL target language; we plan to make that adapotation, as well as our implementation freely navasilable. We also discuss the occurrence of syntactic errors in some of the implementations used for comparison; note on some specifics of neural network settings; and finally comment on comparisons of our work with other different approaches.

The metrics used for comparison in the experiments are \textit{accuracy} (the reason between the number of correctly translated queries and the total of queries) and \textit{syntactical errors} (the reason between the number of queries with syntactical errors and the total of queries).

\subsection{Datasets}

Our experiments were conducted using two traditional datasets: 
\begin{itemize}
\setlength\itemsep{-1ex}
\item \textsf{Geo880}, a set of 880 queries to a database of U.S. geography. The data were originally annotated with Prolog style semantics as the target language, which we manually converted to equivalent statements in SPARQL. On the official web page of \textsf{Geo880} dataset at the University of Austin in Texas one can find a database consisting of Prolog facts,  \textbf{geobase}, and natural language  questions directed at this domain, \textbf{geoquery880}. First we created an OWL ontology based on \textbf{geobase} and then, for each query in natural language in \textbf{geoquery880}, we wrote a corresponding SPARQL query. 

\item \textsf{Jobs640},  a set of 640 queries to a database of job listings. This dataset contains computer-related job postings, such as job announcements, from the USENET newsgroup austin.jobs. We created an OWL ontology based on \textsf{jobs640} and then we wrote a corresponding SPARQL query to each query.
\end{itemize}
Previous work \cite{tang2001using,zettlemoyer2005learning,dong2016language} described on these data sets. Both the ontology and the set of questions can be found in our repository \url{https://github.com/enc-cfgdec/evaluation-data} .

\subsection{Syntactical Errors}

For the purpose of these experiments, a generated SPARQL statement contains a \emph{syntactical error} if it cannot be processed by SPARQL interpreter like Prot\'eg\'e~\footnote{\url{http://protege.stanford.edu/}} due to syntactical formation\footnote{Another option would be to use a grammar that describes SPARQL, then use it as parameter of the parser algorithm. However a rejection by the SPARQL interpreter is a cheaper option found.}. In general, there are several types of syntactical errors that can be generated, such as not closing an open brackets or even a type mismatch between a function and a variable provided as argument.

\subsection{Settings}

Finding better parameters and hyper-parameters for neural networks is always a very costly task. The first rounds with the neural network (pre-tests) served to find the best parameters for the neural network. 
\begin{itemize}
\setlength\itemsep{-1ex}
\item Learning rate: During the pre-test, we use two different learning rates. In the first one, we maintained the learning rate at 0.5. In the second one  has a varying learning rate,  starting at 1 decreasing 5\% at each iteration (epoch). The second case provided the best results;
\item Epoch: In the tests we chose to use 100 epochs, as this was the best result found in our pre-tests for hyperparameters;
\item Hidden dimension: With regard to the number of hidden layers, we used a series of pre-tests with 100, 200 and 400 hidden layers. We continued testing with the 4 different dimensions;
\item Input dimension: We used three different input dimensions, 100, 200 and 300 but the best results were obtained with vectors of size 300.
\end{itemize}

\subsection{Results}

In the experiments, we used 10-fold crossvalidation. In
Table~\ref{tab:res1}, we make a comparison using different $\chi$ context sizes (see Algorithm~\ref{alg:cfgp}).
\begin{table}[H]
\begin{center}
\begin{scriptsize}
\begin{tabular}{c|c|c|c|c}
\cline{1-4}
\multicolumn{1}{ |l|}{\multirow{1}{*}{context size} } & geoquery & jobs640 & train time\\ \cline{1-4}
\multicolumn{1}{ |l|}{\multirow{1}{*}{3} } &
76.53 & 78.36 & 4h &  \\\cline{1-4}
\multicolumn{1}{ |l| }{\multirow{1}{*}{5} } &
79.82 & 81.92 & 6h & \\ \cline{1-4}
\multicolumn{1}{ |l| }{\multirow{1}{*}{$\infty$}} &
83.25 & 85.71 & 11h & \\ \cline{1-4}
\end{tabular}
\caption{Encoder-CFG Decoder with different context sizes}
\label{tab:res1}
\end{scriptsize}
\end{center}
\end{table}

We can see, by Table~\ref{tab:res1}, that the best results were obtained considering $|\chi|=+\infty$. We use this result to make a comparison with our implementation of the know LSTM Encoder-Decoder models \cite{Bahdanau2015} used in \cite{dong2016language} and more simple version with out neural attention.
\begin{table}[H]
\begin{center}
\begin{scriptsize}
\begin{tabular}{c|c|c|c|c|c}
\cline{2-5}
& \multicolumn{2}{ |c| }{Geoquery}& \multicolumn{2}{ c| }{Jobs640} \\ \cline{2-5}
& Accuracy & S.E. & Accuracy & S.E. \\ \cline{1-5}
\multicolumn{1}{ |l|}{\multirow{1}{*}{LSTM Encoder-Decoder} } &
 78.40 & 5.68 & 80.02 & 4.79 &   \\\cline{1-5}
\multicolumn{1}{ |l| }{\multirow{1}{*}{LSTM Encoder-Decoder with Attention} } &
 82.31 & 3.84 & 84.62 & 2.94 &\\ \cline{1-5}
\multicolumn{1}{ |l| }{\multirow{1}{*}{LSTM Encoder CFG-Decoder} } &
83.25 & 0 & 85.71 & 0 &\\ \cline{1-5}
\end{tabular}
\caption{Natural Language to SPARQL comparisons}
\label{tabsparql1}
\end{scriptsize}
\end{center}
\end{table}

Regarding the task of transforming the natural language into SPARQL, we compared our work with \cite{alagha2015using} and \cite{kaufmann2006querix}. In the first paper, the authors also use the Geo880 dataset and through linguistic analysis identify elements of natural language and generate RDF triples. In the second paper, the main strategy of the authors was to try to associate triples of natural language with RDF triples. As can be seen in the Table~\ref{tabsparql2}, we obtained better results in the tests with dataset Geoquery.
\begin{table}[H]
  \begin{center}
    \begin{scriptsize}
      \begin{tabular}{c|r|r|r|}
        \cline{2-3}
        & Geoquery  & Jobs640 \\ \cline{1-3}
        \multicolumn{1}{ |l| }{ \cite{alagha2015using}} & 58.61 & - \\ \cline{1-3}
        \multicolumn{1}{ |l| }{ Querix \cite{kaufmann2006querix} } & 77.67 & - \\ \cline{1-3}
        \multicolumn{1}{ |l| }{LSTM Encoder-Decoder with Attention} & 82.31  & 84.62 \\ \cline{1-3}
        \multicolumn{1}{ |l| }{\bf LSTM Encoder CFG-Decoder} & \bf 83.25  & \bf 85.71 \\ \cline{1-3}
      \end{tabular}
      \caption{Natural Language to SPARQL comparisons}
      \label{tabsparql2}
    \end{scriptsize}
  \end{center}
\end{table}

It is worth mentioning that
in~\cite{tang2001using,zettlemoyer2005learning,dong2016language}, a
kind of logical form of prolog was used as target language, whereas in
out work we consider SPARQL as target language. From the machine
learning point of view, the complexity of the target language is
relevant. For instance, in Geo880 dataset, the pair sentence-logical
form $\{\tuple{x,y}\}$ is 

\begin{small}
\begin{verbatim}
x: parse([what,is,the,density,of,texas,?])
y: answer(A,(density(B,A),const(B,stateid(texas))))
\end{verbatim}
\end{small}
On the other hand, in our SPARQL version we have the pairs
\begin{small}
\begin{verbatim}
x: what is the density of texas ?
y: SELECT((?pop/?area) AS ?density) {?texas p:pop ?pop.?texas p:area ?area}
\end{verbatim}
\end{small}

Besides the complexity, built-in functions such as \verb|density| also simplifies the Geo880 format. These differences give spaces for discussions and questions around comparisons between the approaches.

\section{Discussion and Future Work}
\label{sec:discussion}

The main contribution we give in the present work is the guarantee of translating natural language into a SPARQL query that is gramatically correct. The results also show that the method we propose is competitive with others from literature, therefore we give a relevant contribution to natural language translation.

The main points we observed during the developement of this work and the execution of the experiments lead us to some conclusions: The model learning is more effective, structured and easy in the same proportion that the SPARQL corpus is structured and clear. It means that filters and other complex clauses must be added a posteriori; Althout we did not put any restriction on the CFG, more complex ones may affect the training performance; The proposed training model is at least one and a half times slower than the traditional encoder-decoder model.

The form of translation developed in this work is very important for many other tasks. The fact of obtaining robustness in the target language syntax gives us more confiability to the translation problem where one does not care with the source grammar, but with the target one.

As future work, one may employ mechanisms to synthetize the input of the encoder, in other words, instead of passing the input sentence encoded fixed-length vector representation $c$, a specific $c_i$ that is directly related with the rule to be learned would be passed; this is similar to the strategy adopted in neural attention. Another future work consists in experimenting our model with other versions of CFG.

%

\bibliographystyle{acl}
\bibliography{coling2018}

\begin{thebibliography}{}

\bibitem[\protect\citename{AlAgha}2015]{alagha2015using}
Iyad AlAgha.
\newblock 2015.
\newblock Using linguistic analysis to translate arabic natural language
  queries to {SPARQL}.
\newblock {\em CoRR}, abs/1508.01447.

\bibitem[\protect\citename{Alshawi \bgroup et al.\egroup
  }2014]{alshawi2014deterministic}
Hiyan Alshawi, Pi-Chuan Chang, and Michael Ringgaard.
\newblock 2014.
\newblock Deterministic statistical mapping of sentences to underspecified
  semantics.
\newblock In {\em Computing Meaning}, pages 13--25. Springer.

\bibitem[\protect\citename{Auer \bgroup et al.\egroup }2007]{auer2007dbpedia}
S{\"o}ren Auer, Christian Bizer, Georgi Kobilarov, Jens Lehmann, Richard
  Cyganiak, and Zachary Ives.
\newblock 2007.
\newblock Dbpedia: A nucleus for a web of open data.
\newblock {\em The semantic web}, pages 722--735.

\bibitem[\protect\citename{Bahdanau \bgroup et al.\egroup }2015]{Bahdanau2015}
Dzmitry Bahdanau, Kyunghyun Cho, and Yoshua Bengio.
\newblock 2015.
\newblock Neural machine translation by jointly learning to align and
  translate.
\newblock {\em In Proceedings of the ICLR}.

\bibitem[\protect\citename{Bechhofer \bgroup et al.\egroup
  }2004]{bechhofer2004}
S.~Bechhofer, F.~Van~Harmelen, J.~Hendler, I.~Horrocks, D.L. McGuinness, P.F.
  Patel-Schneider, L.A. Stein, et~al.
\newblock 2004.
\newblock {OWL} web ontology language reference.
\newblock {\em W3C recommendation}, 10:2006--01.

\bibitem[\protect\citename{Bengio \bgroup et al.\egroup
  }1994]{bengio1994learning}
Yoshua Bengio, Patrice Simard, and Paolo Frasconi.
\newblock 1994.
\newblock Learning long-term dependencies with gradient descent is difficult.
\newblock {\em IEEE transactions on neural networks}, 5(2):157--166.

\bibitem[\protect\citename{Borst}1997]{borst1997}
W.N. Borst.
\newblock 1997.
\newblock {\em Construction of engineering ontologies for knowledge sharing and
  reuse}.
\newblock {Ph.D.} thesis, Centre for Telematics and Information Technology
  University of Twente.

\bibitem[\protect\citename{Bowman \bgroup et al.\egroup
  }2014]{bowman2014recursive}
Samuel~R Bowman, Christopher Potts, and Christopher~D Manning.
\newblock 2014.
\newblock Recursive neural networks for learning logical semantics.
\newblock {\em CoRR, abs/1406.1827}, 5.

\bibitem[\protect\citename{Cho \bgroup et al.\egroup }2014]{cho2014learning}
Kyunghyun Cho, Bart Van~Merri{\"e}nboer, Caglar Gulcehre, Dzmitry Bahdanau,
  Fethi Bougares, Holger Schwenk, and Yoshua Bengio.
\newblock 2014.
\newblock Learning phrase representations using rnn encoder-decoder for
  statistical machine translation.
\newblock {\em In Proceedings of the 2014 EMNLP}, pages 1724--1734.

\bibitem[\protect\citename{Dong and Lapata}2016]{dong2016language}
Li~Dong and Mirella Lapata.
\newblock 2016.
\newblock Language to logical form with neural attention.
\newblock In {\em Proceedings of the 54th Annual Meeting of the Association for
  Computational Linguistics (Volume 1: Long Papers)}, volume~1, pages 33--43.

\bibitem[\protect\citename{Earley}1970]{earley1970efficient}
Jay Earley.
\newblock 1970.
\newblock An efficient context-free parsing algorithm.
\newblock {\em Communications of the ACM}, 13(2):94--102.

\bibitem[\protect\citename{Gearon \bgroup et al.\egroup
  }2012]{gearon2012sparql}
Paul Gearon, Alexandre Passant, and Axel Polleres.
\newblock 2012.
\newblock Sparql 1.1 update.
\newblock {\em Working draft WD-sparql11-update-20110512, W3C (May 2011)}.

\bibitem[\protect\citename{Gruber and others}1993]{gruber1993}
T.R. Gruber et~al.
\newblock 1993.
\newblock A translation approach to portable ontology specifications.
\newblock {\em Knowledge acquisition}, 5(2):199--220.

\bibitem[\protect\citename{Hochreiter and Schmidhuber}1997]{hochreiter1997long}
Sepp Hochreiter and J{\"u}rgen Schmidhuber.
\newblock 1997.
\newblock Long short-term memory.
\newblock {\em Neural computation}, 9(8):1735--1780.

\bibitem[\protect\citename{Kaufmann \bgroup et al.\egroup
  }2006]{kaufmann2006querix}
Esther Kaufmann, Abraham Bernstein, and Renato Zumstein.
\newblock 2006.
\newblock Querix: A natural language interface to query ontologies based on
  clarification dialogs.
\newblock In {\em 5th International Semantic Web Conference (ISWC 2006)}, pages
  980--981. Springer.

\bibitem[\protect\citename{Klyne and Carroll}2006]{klyne2006resource}
Graham Klyne and Jeremy~J Carroll.
\newblock 2006.
\newblock Resource description framework (rdf): Concepts and abstract syntax.
\newblock {\em W3C recommendation}.

\bibitem[\protect\citename{Prud’Hommeaux \bgroup et al.\egroup
  }2008]{prud2008sparql}
Eric Prud’Hommeaux, Andy Seaborne, et~al.
\newblock 2008.
\newblock Sparql query language for rdf.
\newblock {\em W3C recommendation}, 15.

\bibitem[\protect\citename{Soru \bgroup et al.\egroup }2017]{soru2017sparql}
Tommaso Soru, Edgard Marx, Diego Moussallem, Gustavo Publio, Andr{\'e}
  Valdestilhas, Diego Esteves, and Ciro~Baron Neto.
\newblock 2017.
\newblock Sparql as a foreign language.
\newblock {\em arXiv preprint arXiv:1708.07624}.

\bibitem[\protect\citename{Tang and Mooney}2001]{tang2001using}
Lappoon~R Tang and Raymond~J Mooney.
\newblock 2001.
\newblock Using multiple clause constructors in inductive logic programming for
  semantic parsing.
\newblock In {\em European Conference on Machine Learning}, pages 466--477.
  Springer.

\bibitem[\protect\citename{Zettlemoyer and
  Collins}2005]{zettlemoyer2005learning}
Luke~S Zettlemoyer and Michael Collins.
\newblock 2005.
\newblock Learning to map sentences to logical form: structured classification
  with probabilistic categorial grammars.
\newblock In {\em Proceedings of the Twenty-First Conference on Uncertainty in
  Artificial Intelligence}, pages 658--666. AUAI Press.

\end{thebibliography}

\end{document}